\pdfoutput=1
\documentclass[11pt]{article}

\usepackage[final]{emnlp2021}

\usepackage{times}
\usepackage{latexsym}
\usepackage{amsmath}
\usepackage{amssymb}
\usepackage{graphicx}
\usepackage{adjustbox}
\usepackage{times}
\usepackage{latexsym}
\usepackage{float}
\usepackage{algorithm}
\usepackage{algpseudocode}
\usepackage{adjustbox}
\usepackage{enumitem}
\usepackage{textcomp}
\usepackage{booktabs}
\usepackage{array}
\usepackage{bbm}
\usepackage{url}
\usepackage{svg}
\usepackage{xcolor}
\usepackage{soul}
\usepackage{seqsplit}
\usepackage{multirow}
\usepackage{adjustbox}
\usepackage{makecell}
\usepackage{xcolor}

\usepackage[T1]{fontenc}
\usepackage[utf8]{inputenc}

\usepackage{microtype}

\newcommand{\name}[0]{Phrase-BERT}
\newcommand\pntm[0]{\textsc{pntm}}
\newcommand{\bmat}[1]{\text{\textbf{#1}}}
\newcommand{\bvec}[1]{\boldsymbol{#1}}

\title{Phrase-BERT: Improved Phrase Embeddings from BERT with an Application to Corpus Exploration}

\author{Shufan Wang \and Laure Thompson \and Mohit Iyyer \\
        University of Massachusetts, Amherst \\
        \texttt{ \{shufanwang, laurejt, miyyer\}@cs.umass.edu} }
\begin{document}
\maketitle
\begin{abstract}
% Despite the success of BERT for many NLP  tasks, is ineffective at representing phrase-level semantic compositionality.
Phrase representations derived from BERT often do not exhibit complex phrasal compositionality, as the model relies instead on lexical similarity to determine semantic relatedness. 
In this paper, we propose a contrastive fine-tuning objective that enables BERT to produce more powerful phrase embeddings. Our approach (\name) relies on a dataset of diverse phrasal paraphrases, which is automatically generated using a paraphrase generation model, as well as a large-scale dataset of phrases in context mined from the Books3 corpus.
\name\ outperforms baselines across a variety of phrase-level similarity tasks, while also demonstrating increased lexical diversity between nearest neighbors in the vector space. Finally, as a case study, we show that \name\ embeddings can be easily integrated with a simple autoencoder to build a phrase-based neural topic model that interprets topics as mixtures of words and phrases by performing a nearest neighbor search in the embedding space.
Crowdsourced evaluations demonstrate that this phrase-based topic model produces more coherent and meaningful topics than baseline word and phrase-level topic models, further validating the utility of \name.

% Finally, as a case study, we integrate our phrase embeddings into a simple neural topic model that represents topics as mixtures of words, phrases, and even sentences.
% Lastly, we conduct a case study on phrase-based topic models to show the semantically meaningful and lexically diverse phrase embeddings derived can be easily integrated into phrase-based neural topic models. Human evaluations show that the derived topic models can generate coherent and meaningful topic descriptions using not just unigrams but diverse lexical units such as phrases and even short sentences.

\end{abstract}

% In this work, we evaluate the phrase representations from BERT on a variety of tasks, relying on human judgements on phrase semantics. 
% We also propose a simple fine-tuning strategy to derive semantically meaningful phrase embeddings from BERT, by using a constrastive learning objective, with automatically generated phrase-level paraphrases, as well as longer masked contexts.
% We show that the derived phrase embeddings not only lead to better performance on all phrase semantics evaluation tasks, but also yield more lexical diversity in the nearest neighbours, as the derived phrase embeddings rely less on lexical overlap to determine semantic similarity.
\section{Introduction} 
Learning representations of phrases is important for many tasks, such as semantic parsing \citep{socher-2011-phrase-rnn}, machine translation \citep{Ramisch-phrase-important}, and question answering \citep{seo-etal-2018-phrase-important-qa}.
While pretrained language models such as BERT \citep{devlin2018bert} have significantly pushed forward the state of the art on a variety of NLP tasks, they still struggle to produce semantically meaningful embeddings for shorter linguistic units such as sentences and phrases. An out-of-the-box BERT sentence embedding model often underperforms simple baselines such as averaging GloVe vectors in semantic textual similarity tasks \citep{reimers-2019-sentence-bert}.
% which are not contextualized representations and trained with simpler architecture. 
Furthermore, \citet{yu2020assessing} have shown that phrasal representations derived from BERT do not exhibit complex phrasal compositionality.

\begin{table}[t]
	\begin{center}
		\footnotesize
		\begin{tabular}{ p{1cm}p{5.5cm} }
		\toprule
		\bf  Model & \bf Nearest neighbors of \textbf{``pulls the trigger"} \\
		\midrule
		GloVe & his \textcolor{red}{trigger}, the \textcolor{red}{trigger}, a \textcolor{red}{trigger}\\
% 		\textcolor{red}{pulling} his \textcolor{red}{trigger}\\	
		\midrule

		BERT & \textcolor{red}{pulled} the \textcolor{red}{trigger}, squeezed the \textcolor{red}{trigger}, scoots closer\\
% 		\textcolor{red}{pulling} the \textcolor{red}{trigger} \\
		\midrule

	    Span-BERT & \textcolor{red}{pulled} the \textcolor{red}{trigger}, \textcolor{red}{pulling} the \textcolor{red}{trigger}, seize the day\\
	   % hanging in the balance  \\
		\midrule

		Sent-BERT & \textcolor{red}{pulling} the \textcolor{red}{trigger}, \textcolor{red}{pulled} the \textcolor{red}{trigger}, the \textcolor{red}{trigger}\\
% 		squeezed the \textcolor{red}{trigger}\\
		\midrule

		\name & \textbf{picks up his gun, squeezes off a quick burst of shots, takes aim}\\
% 		makes a move}\\ 
		
		\bottomrule
		\end{tabular}
\end{center}
\caption{Nearest neighbors of the phrase ``pulls the trigger''. While baselines rely heavily on  \textcolor{red}{lexical overlap}, \name's neighbors are both semantically similar and lexically diverse.}
\label{tab:nn}
\end{table}

In this paper, we develop \name, which fine-tunes BERT  using contrastive learning to induce more powerful phrase embeddings. Our approach directly targets two major weaknesses of out-of-the-box BERT phrase embeddings: (1) BERT never sees short texts (e.g., phrases) during pretraining, as its inputs are chunks of 512 tokens; and (2) BERT relies heavily on lexical similarity (word overlap) between input texts to determine semantic relatedness \citep{li2020-bertflow, yu2020assessing, paws2019naacl}.
To combat these issues, we automatically generate a dataset of lexically-diverse phrasal paraphrase pairs, and we additionally extract a large-scale dataset of 300 million phrases in context from the Books3 dataset from the Pile \citep{pile}. We then use this paraphrase data and contextual information to fine-tune BERT with an objective that intuitively places phrase embeddings close to both their paraphrases and the contexts in which they appear (Figure \ref{fig:emb-space}).

% \swcomment{added: One main challenge is that sent labelled data and sent eval is very rich, not so much for phrase, therefore, we generate
% }

\name\ outperforms strong baselines such as SpanBERT \citep{joshi2019spanbert} and Sentence-BERT \citep{reimers-2019-sentence-bert} across a suite of phrase-level semantic relatedness tasks. Additionally, we show that its nearest neighbor space exhibits increased lexical diversity, which signals that compositionality plays a larger role in its vector space (Table \ref{tab:nn}). Such phrasal diversity is an important component of models built for corpus exploration such as phrase-based topic modeling \citep{wang2007topical,griffiths2007topics}. To investigate \name's potential role in such applications, we integrate it into a neural topic model 
% This diversity is extremely useful for corpus exploration \citep{ hideki-phrase-div-important, schofield-mimno-2016-phrase-div-important, melucci-phrase-div-important, yass-phrase-div-important, dieng2019-phrase-div-important-topic-model,
% bianchi-bert-topic-model} \micomment{cite something if you can find it} \ltcomment{maybe i missed something, but what does stemming in this case have to do with phrase/lexical diversity? might be worth splitting into two sentences (1) on literature (2) on our model}, 
that represents topics as mixtures of words, phrases, and even sentences. A series of human evaluations reveals that our phrase-level topic model produces more meaningful and coherent topics compared to baseline models such as LDA \citep{blei2003latent} and its phrase-augmented variants. We have publicly released code and pretrained models for \name\ to spur future research on phrase-based NLP tasks.\footnote{ \href{ https://github.com/sf-wa-326/phrase-bert-topic-model }{ https://github.com/sf-wa-326/phrase-bert-topic-model } }

\section{Related work}
\label{sec:related}
Our work relates to a long history of learning dense phrase representations, and in particular to approaches that leverage large-scale pretrained language models. Like most prior approaches, \name\ learns a \emph{composition} function that combines component word embeddings together into a single phrase embedding. This function has previously been implemented with rule-based composition over word vectors \citep{yu-dredze-2015-phrase-emb-compos} and recurrent models \citep{zhou-2017-phrase-emb-gru} that use a pair-wise GRU model using datasets such as PPDB \citep{pavlick2015ppdb}. Other work learns \emph{task-specific} phrase embeddings such as those for semantic parsing \citep{socher-2011-phrase-rnn}, machine translation \cite{bing-etal-2015-phrase-important-summarization} and question answering \citep{dense-phrase-qa}; in contrast, \name\ produces general-purpose embeddings useful for any task. 

The advent of huge-scale pretrained language models such as BERT \citep{devlin2018bert} has opened a new direction of phrase representation learning. \citet{yu2020assessing} highlight BERT's struggles to meaningfully represent short linguistic units (words, phrases). Several papers hypothesize that this is because BERT is trained on longer texts (512 tokens) and with a pairwise text objective that may be irrelevant for shorter texts \citep{reimers-2019-sentence-bert,Liu2019RoBERTaAR,toshniwal-etal-2020-cross}. Without task-specific fine-tuning, the performance of BERT on phrases and sentences is often worse than simple baselines such as mean-pooling over GloVe vectors \citep{reimers-2019-sentence-bert, li2020-bertflow}. Furthermore, \citet{li2020-bertflow} draw theoretical connections between BERT's pretraining objective and its non-smooth anisotropic semantic embedding space, which make it more reliant on lexical overlap to determine phrase and sentence similarity. Previously proposed methods to address these issues include predicting spans during pretraining instead of words \citep{joshi2019spanbert},  fine-tuning BERT on shorter texts \citep{reimers-2019-sentence-bert}, and adding an explicit postprocessing step to induce a continuous and isotropic semantic space \cite{li2020-bertflow}. As we show in the rest of this paper, \name\ produces more semantically meaningful phrase representations than these competing approaches while also promoting a lexically diverse vector space.

\section{Phrase Embeddings from BERT}
\label{sec:method}

\begin{figure}[t]
 \centering
  \includegraphics[height=0.6\linewidth, width=0.9\linewidth]{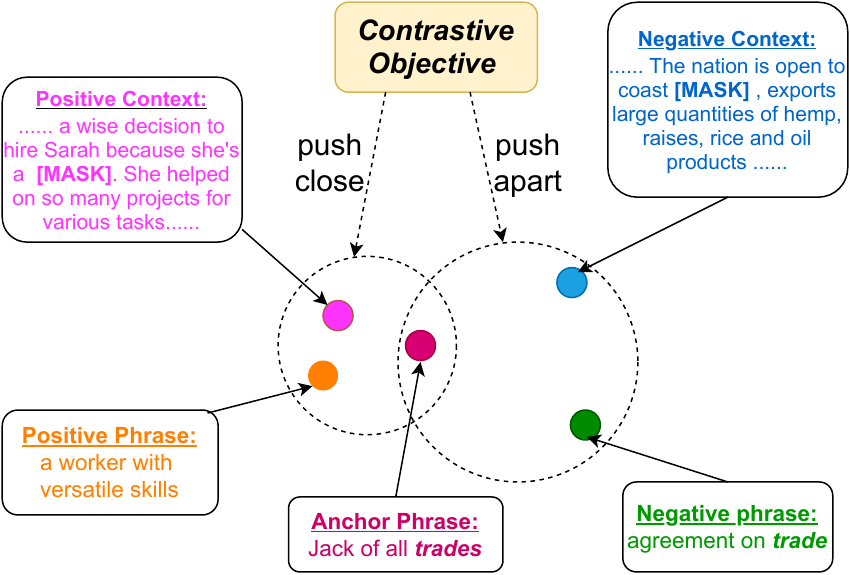}
  \caption{We propose \name\ to correct BERT's embedding space for phrases by placing semantically similar phrases and contexts closer together (e.g., the anchor phrase ``Jack of all trade'' and the positive paraphrase ``a worker with versatile skills'') and by removing lexical cues with a contrastive learning objective.}
  \label{fig:emb-space}
\end{figure}

We design two separate fine-tuning tasks on top of BERT to improve its ability to produce meaningful phrase embeddings. Since the pretrained BERT model relies heavily on lexical overlap to determine phrase similarity, our first fine-tuning objective relies on an automatically generated dataset of \textbf{lexically diverse phrasal paraphrases} to encourage the model to move beyond string matching. The second objective encourages the model to encode contextual information into phrase embeddings by relying on a phrase-in-context dataset we extract of \textbf{phrases in context} from the huge-scale Books3 corpus \citep{pile}. In both cases, we rely on contrastive objectives similar to Sentence-BERT \citep{reimers-2019-sentence-bert} for fine-tuning (Fig. \ref{fig:emb-space}).

\paragraph{Using BERT to embed phrases:} Given an input phrase $X$ of length $N$ tokens, we compute a  representation $\bvec{x}$ by averaging the final-layer token-level vectors yielded by BERT \citep{devlin2018bert}\footnote{We use the 12-layer BERT-base-uncased for all experiments.} after passing the tokens of $X$ to the model as input: $
    \bvec{x} = \sum_{i=1}^{N} \textsc{bert}(X_i) / N. $
As all of BERT's pretraining examples are 512 tokens long, the above method is reliable for short documents, but it struggles to model the semantics of words and phrases, as shown by \citet{yu2020assessing} and also by our evaluations in Section \ref{sec:phrase-eval-tasks}.

% After sentence-level fine-tuning, the resulting model places phrases with high lexical overlap near each other even if they are semantically dissimilar.
% Thus, we fine-tune this model on an automatically-generated dataset of phrase-level paraphrase pairs, which is specifically constructed to minimize lexical overlap.\footnote{In preliminary experiments, we also tried using PPDB~\citep{pavlick2015ppdb} instead of creating our own data, but it has low phrase coverage and many paraphrases are surface variations with minimal lexical difference.} This step forms our work's novel contribution and is crucial to improving the quality of our topic model; as shown in the last row of Table~\ref{tab:nn}, it results in topically coherent neighbors such as ``is located in the southwestern united states'' and ``arizona territory'' with minimal lexical overlap. 

\paragraph{Creating lexically diverse phrase-level paraphrases:}
Our first fine-tuning objective encourages BERT to capture semantic relatedness between phrases without overly relying on lexical similarity  between those phrases.
To accomplish this, we create a dataset by extracting 100K phrases from WikiText-103 \citep{wk103} using the shift-reduce parser from CoreNLP \citep{corenlp}.\footnote{We extract all NP, VP, ADJP, and ADVP phrases and then filter to select the most frequent 100K. More details on this process can be found in the Appendix \S \ref{sec:source_phrase_extraction}.}
Then, given a phrase $p$ ``complete control'' from the sentence ``The local authorities have complete control over the allocation of building materials'', we create a positive example $p^+$ by passing $p$ through the GPT2-based \emph{diverse} paraphrasing model released by \citet{style20}. This model was trained by fine-tuning GPT2-large \cite{Radford2019LanguageMA} on a filtered version of the PARANMT-50M sentence-level paraphrase dataset~\citep{paranmt50}, using an encoder-free seq2seq modeling approach as proposed by \citet{wolf18}. 

We decode from this model using nucleus sampling with  the nucleus probability mass of $0.8$ \citep{holtzman2019curious}, applying lexical constraints to avoid producing any non-stopword tokens that also occur in $p$. This yields phrases such as ``full power of the system'', which are quasi-paraphrases of $p$ with no lexical overlap.
We create a negative example $p^-$ by first randomly sampling a non-stopword from $p$ and then replacing it with a random token from the vocabulary. In the case of ``complete control'', we might sample ``complete'' and replace it with a randomly selected token ``fluid''. Then, we feed the corrupted phrase into the paraphraser and decode just as we did to produce the positive example, which removes lexical overlap but preserves the distorted meaning. This produces phrases like ``no change to the water level'', which has no semantic relation to $p$.

\paragraph{Collecting phrases in context:}
The above dataset focuses exclusively on phrases \emph{out of context}. In other words, a model trained to distinguish negative phrases from positive phrases does not observe any surrounding context in which these phrases are used. As these contexts also provide useful information about the meaning and usage of phrases, we create a second dataset to inject contextual information into BERT's phrase embeddings. Concretely, we extract phrases along with their surrounding context from the Books3 Corpus \citep{pile}, a large-scale 100GB collection of books spanning a variety of topics and genres. As before, we extract phrases by running constituency parsing on a random subset of the dataset; we remove all phrases that are more than ten tokens long and then select the top 100K most frequent phrases. We also store a single positive context $c^+$ of length 120 tokens in which $p$ occurs, replacing the occurrence of $p$ within $c^+$ with a \texttt{MASK} token.
% The phrase-context dataset includes text from far more domains and also introduces length variability (the context text can be as long as 128 tokens) and is therefore critical in introducing the desired lexical diversity into the nearest neighbours of \name's embedding space, as we will show in our diversity evaluation in \ref{sec:div}.

\subsection{Fine-tuning BERT with a contrastive objective using the constructed datasets }

We fine-tune BERT on both datasets with the same contrastive objective, following similar procedures as Sentence-BERT \citep{reimers-2019-sentence-bert}. For the first dataset, we encourage the model to produce similar embeddings for $p$ and $p^+$ while pushing the embeddings for $p$ and $p^-$ far apart. We embed each phrase in the triplet ($p$, $p^+$, $p^-$) by mean-pooling BERT's token-level representations as described previously, which gives us three vectors $(\bvec{p}, \bvec{p}^+, \bvec{p}^-)$ Then, we compute the following triplet loss:

\begin{equation} J = \max(
0, \epsilon
- \lVert \bvec{p} - \bvec{p}^-\rVert  
+ \lVert \bvec{p} - \bvec{p}^+\rVert
)
\end{equation}
where  $\lVert \cdot \rVert $ denotes the L2 norm and $\epsilon$ is a margin (set to 1 in our experiments).

Similarly, for the second dataset, we compute the triplet loss,
% \begin{equation} 
$\max(
0, \epsilon
- \lVert \bvec{p} - \bvec{c}^-\rVert  
+ \lVert \bvec{p} - \bvec{c}^+\rVert
)$,
% \end{equation}
for each data instance $(\bvec{p}, \bvec{c}^+, \bvec{c}^-)$, or embedding vectors encoded by \name\ from the phrase-context triple ($p$, $c^+$, $c^-$), where $c^-$ is a randomly sampled context. 

\paragraph{Implementation details: } We fine-tune Phrase-BERT on an NVIDIA RTX 2080ti GPU for 1 epoch. We use a batch size of 16 and optimize using Adam~\citep{kingma2014adam} with a learning rate of $2e-5$. The initial $10\%$ of training steps are used as warm-up steps, following the linear warm-up schedule used by \citet{reimers-2019-sentence-bert}.

% \micomment{write eqn for second objective using c+ and c- instead of p. also explain how you encode the contexts} \swcomment{repetitive to write the same equation twice, move to footnote?}

\section{Experimental setup}
We evaluate our phrase embeddings on a diverse collection of phrase-level semantic relatedness tasks following previous works on evaluating phrase embeddings \citep{Turney, yu-dredze-2015-phrase-emb-compos,asaadi-etal-2019-bird, yu2020assessing}.
% (Table~\ref{tab:phraseeval}) derived from existing datasets such as PPDB~\citep{pavlick2015ppdb} and PAWS~\citep{paws2019naacl}. 
Due to a lack of benchmarks like SentEval \citep{conneau2018-sent-eval} at the phrase level, we create filtered versions of some datasets by removing lexical overlap cues.

% Unlike evaluation for sentence representations such as , there is no standard phrase semantics evaluation benchmark. Previous works on phrase embeddings use methods such as phrase-unigram match, correlation with human judgements and paraphrase identification to measure the quality of phrase embeddings to capture phrase semantics. However, many such tasks only involve phrase of similar length (mostly bigrams). We present a collection of semantics evaluation tasks, to assess \name's ability to encode meanings of phrases with a variety of length. We also perform filtering on some the existing datasets to remove cues from word overlap, so as to test the models ability to capture complex meaning composition, beyond simply using lexical overlap information. 

\subsection{Datasets}
\label{sec:phrase-eval-tasks}
% We compare the performance of \name\ against baselines on the following tasks. Examples and details of each are provided in .

We compare the performance of \name\ against baselines on a variety of phrases tasks involving phrases of different length and types.

\paragraph{Turney:} The dataset of \citet{Turney} contains 2,180 examples that test bigram compositionality by asking models to select which of five \emph{unigrams} is most similar in meaning to a given bigram.

\paragraph{BiRD:} The bigram-relatedness judgment dataset \citep{asaadi-etal-2019-bird} is a correlation task that consists of 3,455 pairs of bigram phrases, each of which has a corresponding human rating of similarity between 0 and 1. 

\paragraph{PPDB:} We create a paraphrase classification dataset from PPDB 2.0 \citep{pavlick2015ppdb} that contains 23,364 phrase pairs\footnote{We use a 70/15/15 train/dev/test split.} by sampling examples from PPDB-small, the highest-quality subset of PPDB 2.0 according to human raters. Given a pair, we apply the paraphrase classification model described later in this section to input phrase embeddings to determine if the inputs are paraphrases. Negative examples are created by randomly sampling phrases from the dataset. The average phrase length in this dataset is 2.5 tokens.

\paragraph{PPDB-filtered:} Since the above PPDB dataset contains a large amount of lexical overlap between paraphrase pairs, it can be solved with superficial heuristics. We follow \citet{yu2020assessing} by creating a more challenging version by filtering out lexical overlap cues. Specifically, we control the amount of word overlap in each positive and negative pair to be exactly the same. We also ensure that each overlapping token in a pair occurs in both positive and negative pairs so that the model cannot rely on cues from word identity. This dataset has 19,416 phrase pairs.

% We adopt the method from \citet{yu2020assessing} with additional filtering to gather a paraphrase classification dataset, removing cues from lexical overlap.

\paragraph{PAWS-short:} The previous datasets test include mainly bigrams and short phrases, motivating us to evaluate our models on a dataset with longer text. PAWS is a challenging dataset for paraphrase classification on text pairs where even negatives have high lexical overlap \citep{paws2019naacl}. However, it contains sentences and short paragraphs in addition to phrases. We filter PAWS to only include  examples shorter than 10 tokens in length while ensuring class balance between paraphrase and non-paraphrase pairs. We follow the split released by the authors and extract 1,300 total examples, with an average length of 9.4 tokens.

% We will show in Section \ref{sec:phrase-eval} that baseline models such as BERT and even modified models such as sentence-Bert and spanBERT cannot represent phrase semantics well, indicated by its poor performance across the evaluation tasks. \name\ consistently outperforms baselines across evaluation datasets, especially for the datasets where lexical overlap cues are removed (PPDB-filtered and PAWS-short), suggesting that it is better able to capture complex phrase meaning compositions beyond word content.

\begin{table}
	\begin{center}
		\small
    \begin{adjustbox}{max width=\textwidth}
		\begin{tabular}{ p{1cm}p{4cm}p{1cm}}
            % column names
			\toprule
			\textbf{Dataset} & \textbf{Example} 
			& \textbf{Label}\\
			\midrule
			
			% dataset
			Turney
			& 
			% example
			(learned person, pundit) 
			&
			% label
			``match''
			\\
			\midrule

			% dataset
			BiRD
			& 
			% example
			(business development,\newline economic growth) 
			&
			% label
			$0.586$ 
			\\
			\midrule

			% dataset
            PPDB-filtered
			& 
			% example
			(global \textcolor{red}{affairs}, world \textcolor{red}{affairs}) 
			&
			``positive'' 
			\\
			\cmidrule{2-3}

			% dataset
            % PPDB-filtered
			& 
			% example
			(world \textcolor{red}{affairs}, domestic \textcolor{red}{affairs}) 
			&
			%label
			`negative''
			\\
			\midrule

			% dataset
            PPDB
			& 
			% example
			(actively participate, \newline
			play an activate role)
			&
			``positive"
			\\
			\midrule

			% dataset
            PAWS-short
			& 			
			% example
			(a \textcolor{red}{variable} version of the \textcolor{red}{basic} lyrics, \newline
			a \textcolor{red}{basic} version of the \textcolor{red}{variable} lyrics) 
			&
			``negative"
			\\
			\bottomrule
	
		\end{tabular}
    \end{adjustbox}
	\end{center}
	\caption{Datasets and examples used in our phrase embedding evaluation.}
	\label{tab:dataset}
\end{table}

\subsection{Baselines} We compare \name\ against phrase embeddings derived from baselines that include averaged GloVe vectors\footnote{https://nlp.stanford.edu/projects/glove/} as well as the base versions of BERT \cite{devlin2018bert}, SpanBERT \citep{joshi2019spanbert}, and Sentence-Bert \citep{reimers-2019-sentence-bert}. Except for GloVe and Span-BERT, We obtain phrase embeddings from GloVe by averaging pretrained token embeddings; for Span-BERT, we use the concatenation of the phrase boundary representations following \citet{joshi2019spanbert}. we use the mean-pooled representation over the final-layer outputs from these models as phrase representations, following the observation by \citet{reimers-2019-sentence-bert} and \citet{yu2020assessing} that this method outperforms other possibilities (e.g., using the \textsc{[cls]} representation). We also compare the full \name\ model with two ablated versions: \name-phrase (removing the phrase-context fine-tuning) and \name-context (removing the phrase-level paraphrase fine-tuning).

\paragraph{Paraphrase classification model:} The paraphrase classification datasets (PPDB-filtered, PPDB, and PAWS-short)  require task-specific fine-tuning. We use the same setup as \citet{adi-16} and \citet{yu2020assessing}. In short, we add a simple classifier on top of the concatenated embedding of a phrase pair, implemented using an multilayer perceptron with a hidden layer of size 256 and an ReLu activation before the classification layer.

\begin{table}[t]
    \centering\small
    
    \begin{adjustbox}{max width=0.455\textwidth}

        % \begin{tabular}{crrrrr}
        \begin{tabular}{p{1.8cm}p{0.7cm}p{0.7cm}p{0.7cm}p{0.7cm}p{0.7cm}}
        \toprule
          \bf Model & \bf Turney & \bf BiRD & \bf PPDB-filtered & \bf PPDB & \bf PAWS-short \\
          \midrule
        
          GloVe & 37.8
            & 0.560
            & 44.2
            & 47.2
            & 50.0 \\
            
          BERT &42.6
            & 0.444
            & 60.1
            & 86.2
            & 50.0 \\
            
          SpanBERT & 38.7
            & 0.258
            & 57.3
            & 95.1
            & 50.1 \\
            
          Sentence-BERT & 51.8
            & 0.687
            & 64.2
            & 95.8
            & 50.0 \\
        
          \midrule
          Phrase-BERT-phrase & 55.4
            & 0.682
            & 68.0
            & 96.9
            & 50.0 \\
        
          Phrase-BERT-context & 55.0
            & 0.672
            & 65.0
            & 95.7
            & 49.2 \\
          \midrule
          Phrase-BERT & \textbf{57.2}
            & \textbf{0.688}
            & \textbf{68.0}
            & \textbf{97.6}
            & \textbf{58.9} \\

          \bottomrule
        \end{tabular}
        \end{adjustbox}
    \caption{\name\ outperforms other baselines on all phrase-level semantic relatedness datasets. The improvements on PPDB-filtered and PAWS-short show that \name\ captures phrase semantics without over-reliance on lexical overlap.}
    \label{tab:phraseeval}
\end{table}

\section{Results \& Discussion}
In this section, we highlight takeaways from our results on the phrase-level semantic relatedness benchmarks as well as measurements of lexical diversity. We also provide an ablation study that confirms the benefits of both fine-tuning objectives.

\subsection{\name\ effectively captures phrase semantics }
\label{sec:phrase-eval}
From Table \ref{tab:phraseeval}, we observe that \name\ consistently outperforms BERT and other baseline models across all five evaluation datasets. Among the baselines, Sentence-BERT also yields notable improvements over BERT, demonstrating the relationship between phrase and sentence-level semantics. However, \name\ still outperforms Sentence-BERT, especially in tasks where the input is very short, such as the phrase-unigram tasks from Turney. Moreover, despite the masked span prediction objective of SpanBERT, which intuitively may increase its ability to represent phrases, the model consistently underperforms on all tasks.

\subsection{\name\ does not rely solely on lexical information to understand phrases.} 
Several previous works report that pretrained transformer-based representations overly use lexical overlap to determine semantic relatedness \citep{yu2020assessing, li2020-bertflow, reimers-2019-sentence-bert}. Our experiments quantitatively show that for both short and long phrases, BERT and other baselines heavily rely on lexical overlap and not compositionality to encode phrase relatedness. Despite high accuracies on the full PPDB dataset (where examples with lexical overlap are not filtered out), baselines significantly underperform \name\ on the two datasets in which lexical overlap cues are removed for paraphrase classification (PPDB-filtered, PAWS-short).
On the other hand, \name's strong across-the-board performance demonstrates that it is able to go beyond string matching.
Additionally, both of \name's objectives are complementary: \name-phrase (trained with paraphrase data only) and \name-context (trained with context data) are both consistently worse than \name. 
% Additionally, the context data training is crucial to the model's ability to recognize phrase semantics beyond lexical similarity, which will be discussed in more details in Section~\ref{sec:ablation}.

\subsection{Evaluating lexical diversity in the phrase embedding space}
\label{sec:div}
For many practical use cases of phrase embeddings (e.g, corpus exploration, or tracking how phrasal semantics change over time), it is useful to visualize the nearest neighbors of particular phrases \citep{mikolov-phrase-2013, dieng2019-phrase-div-important-topic-model, bommasani-etal-2020-interpreting}.  However, if these nearest neighbors contain heavy lexical overlap, they may not add any new information and may miss important meaning from phrases. For example, the phrase ``natural language processing'' has no lexical overlap with ``computational linguistics'', but both should be nearest neighbors. To measure this,  we obtain the top-$10$ nearest neighbors for a query phrase in the embedding space and measure the lexical diversity within this set.\footnote{We randomly choose source phrases from the Storium dataset \citep{storium2020}, which contains a diverse set of stories that does not appear in either the pretraining or fine-tuning data of \name, and use a vocabulary of 125K most frequent words and phrases from this dataset to compute the nearest neighbors.} We report three different metrics: (1) the \textbf{percentage of unique unigrams}
in each of the phrase's nearest neighbors normalized by the phrase's length, which is inspired by sentence-level translation diversity metrics \citep{div-frac-2018}; (2) \textbf{LCS-precision}, which measures the longest common substring between the source phrase and the top-$k$ nearest neighbors (lower = more diverse); and (3) the average \textbf{Levenshstein edit distance} \citep{levenshtein1966binary} between a phrase and each of its top-$k$ nearest neighbors.
Table~\ref{tab:phrasediv} shows that \name\ exhibits slightly higher lexical diversity than Sentence-BERT, which is the most competitive model on semantic relatedness tasks after \name. 

\subsection{Ablating the two objectives}
As shown in Table~\ref{tab:phraseeval}, \name-phrase also performs reasonably well in many phrase semantics tasks, especially the PPDB paraphrase classification tasks. However, without training on the context data, which is much longer (128 tokens), it underperforms on the PAWS-short dataset, which consists of longer inputs. \name-phrase is also worse at inducing a lexically diverse embedding space, as indicated by the high LCS-precision. Meanwhile, fine-tuning using only the context objective (\name-context) yields the highest lexical diversity (Table~\ref{tab:phrasediv}) at the cost of a worse semantic space, which is perhaps because of the diverse content in the extracted contexts.  
% However, \name-context does not perform well in the two paraphrase classification tasks compared to \name-phrase, indicating that the the paraphrase data is still needed to build a phrase embedding model that captures nuanced phrase semantics.

% \subsection{Inducing lexical diversity without sacrificing phrase semantics}
% Among the nearest neighbours of phrases induced by all models, \name\ induces high percentage of unique tokens, high edit distance among phrase pairs, as well as low common-substring precision. This show taht it is able to induce high lexical diversity, while retaining the ability to capture phrase semantics. While BERT also gives high diversity measure, this is at the cost of not giving effective phrase representations(indicated by its low performance in every phrase semantics evaluation tasks). Hence the phrase embedding space of BERT is not as meaningful, and high lexical diversity is achieved at the expense of meaning divergence. \name\ is able to improve the lexical diversity, while still preserving semantic meanings in the embedding space. 

\begin{table}[t]
    \centering\small
\begin{adjustbox}{max width=0.475\textwidth}

\begin{tabular}{p{2.9cm}p{1cm}p{1cm}p{1.5cm}}
\toprule
  \bf Model & \bf \% new tokens & \bf LCS-precision & \bf Levenshtein-Distance \\
  \midrule

%   GloVe & 4.1
%     & 53.0
%     & 8.3 \\
    
%   BERT & 5.7
%     & 39.5
%     & 8.8 \\
    
%   spanBERT & 6.7
%     & 35.2
%     & 11.4 \\
    
  Sentence-BERT & 5.0
    & 51.1
    & 8.5 \\
    
  Phrase-BERT & 5.3
    & 47.6
    & 8.7 \\

  \midrule
  Phrase-BERT-phrase & 4.8
    & 52.3
    & 8.2 \\

  Phrase-BERT-context & 5.4
    & 44.8
    & 9.0 \\
    
%   \midrule

  \bottomrule
\end{tabular}
\end{adjustbox}

    \caption{Lexical diversity among the top-10 nearest neighbors in the phrase embedding space.}
    \label{tab:phrasediv}
\end{table}

% \paragraph{Ablation on fine-tuning data}
% \name\ is trained on using the pooled dataset of \textbf{phrase-level paraphrases} and \textbf{phrase-in-contexts}. We conduct ablation studies by training \name-phrases (using only the paraphrases data) and \name-context (using the context data). 

% \subsection{Ablation Study}

% \input{sections/6-experiment}
% \input{sections/7-result}
\section{Using \name\ for topic modeling}

We have shown that \name\ produces meaningful embeddings of variable-length phrases and a lexically diverse nearest neighbor space. In this section, we demonstrate \name's utility in the downstream application of phrase-based corpus exploration.
Capturing both phrase semantics and phrasal diversity is an important aspect for topic models that incorporate phrases in topic descriptions \citep{wang2007topical, griffiths2007topics,topmine}. We show that  \name\ can be integrated with an autoencoder model to build a phrase-based neural topic model (\pntm). Despite its simple architecture, \pntm\ outperforms other topic model baselines in our human evaluation studies in terms of topic coherence and topic-to-document relatedness (Table~\ref{tab:topicdump}). 

\begin{table}[t]
	\begin{center}
		\small
		\begin{tabular}{ p{7cm}}
			\toprule
			
            the high seas fleet,
            wartime,
            kamikaze,
            the imperial japanese navy,
            the outbreak of world war ii,
            guadalcanal\\ \midrule
            
			an award,
            critically acclaimed,
            woman of the year,
            best actor,
            awards and nominations,
            the winner,
            best actress\\ \midrule
            
            hindi,
            the indian ocean,
            subcontinent,
            the central bay of bengal,
            bengali,
            india 's,
            bihar\\

            \midrule
            rhythmic,
            monosyllable,
            beats,
            the song 's composition,
            drumbeat,
            rhythmically,
            the song 's lyrics\\ 
            \midrule
            
            stalking,
            the mystery,
            paranormal,
            linked to the paranormal,
            fox mulder david duchovny,
            cases linked to the paranormal,
            the conspiracy,
            mulder and scully\\ \midrule
            
            a separate species,
            phylogenetic,
            taxonomic,
            clade,
            a genus,
            taxonomical,
            phylogenetically\\
			\bottomrule

		\end{tabular}
	\end{center}
	\caption{A sample of six topics induced by \pntm\ with $K=1000$ on Wikipedia. Topics are interpreted as mixtures of words and phrases, which enables fine-grained exploration of document collections.}
	\label{tab:topicdump}
\end{table}

\subsection{Building a topic model with \name}

We integrate \name\ into previous unigram-based neural topic models \citep{rmn2016, storium2020}
that try to reconstruct a document representation through an interpretable bottleneck layer. 
% Unlike prior implementations, computing text representations using \name\ allows us to meaingfully interpret the topic model's bottleneck parameters with both words and phrases that are semantically coherent and lexically diverse.  
Unlike prior implementations, computing text representations using \name\ allows us to produce high quality topic descriptions (with a mixture of words and phrases) using a simple nearest neighbor search in the embedding space \footnote{We also conduct experiments of training other versions \pntm, replacing the embedding function \name\ by other composing functions such as BERT and SpanBERT. This leads to issues such as incoherence and the lack of lexical diversity in topic descriptions, further highlighting the strength of \name\ in capturing phrase semantics. Examples of topics obtained with these models are provided in Appendix \S \ref{sec:topicdump-qualitative}}.

\subsubsection{Model description}
The bottleneck layer in our \pntm\ is implemented through a linear combination of rows in a $K \times d$ dimensional topic embedding matrix $\bmat{R}$, where $K$ denotes the number of topics, each row of $\bmat{R}$ corresponds to a different topic's embedding and $d=768$. 
Concretely, assume we have an input document $X$ with tokens $X_1, X_2, \dots, X_n$. We encode the document by passing its tokens through \name\ to obtain a single vector representation $\bvec{x}$.
We then score $\bvec{x}$ against $K$ different learned topic embeddings,\footnote{This computation is identical to a dot product attention mechanism \citep{bahdanau2014neural}.} which produces a distribution $t$ over topics: $
    t = \text{softmax}(\bmat{R}\bvec{x}).$
Given the distribution $t$, we then compute a reconstructed vector $\bvec{\tilde{x}}$ as a linear combination of the rows in $\bmat{R}$: $
    \bvec{\tilde{x}} = \bmat{R}^\mathsf{T} t$.

Intuitively, we want the model to push $\bvec{\tilde{x}}$ as close to the input $\bvec{x}$ as possible as this forces salient information to be encoded into the rows of $\bmat{R}$. We accomplish this through optimizing a contrastive loss function, which  minimizes the inner product between $\bvec{\tilde{x}}$ and $\bvec{x}$ while maximizing the inner product between $\bvec{\tilde{x}}$ and the representation $\bvec{z}$ of some randomly sampled document $Z$:
% Given an input document representation $x$, we randomly sample five other documents in the same dataset and compute the average of their representations, $x_n$. Then we minimize the triplet loss function $L(\theta)$, using the correct input representation $x$, the reconstructed representation $\hat{x}$ and the average of randomly sampled representation $x_n$,

\begin{equation} L(\theta) = \sum_{i}^N \max(0, 1 - \bvec{\tilde{x}} \cdot \bvec{x}  + \bvec{x} \cdot \bvec{z}_i ),
\end{equation}

where the summation is over $N$ negative documents ($N$=$5$ in our experiments), and $\theta$ denotes the model parameters. Finally, to discourage duplicate topics, we follow~\citet{rmn2016} by adding an orthogonality penalty term 
    $H(\theta) = \lVert \bmat{R} \bmat{R}^{T} - \bmat{I} \rVert,$
where $\bmat{I}$ is the identity matrix.

% $H$ is also a regularization term that enforces orthogonality on the topic matrix $R$ \citep{ica2000}. Therfore, the final training objective $J$ for a single document $X$ is,

% \begin{equation}
%     J(\theta) = L(\theta) + \lambda H(\theta)
% \end{equation}

% where $\lambda$ is a hyperparameter that controls the diversity of the topics learned.

\subsubsection{Interpreting learned topic embeddings}
\label{sec:intepret-topics}
 After training \pntm\ over all of the documents in a target collection, we obtain topics (lists of words and phrases that are most closely associated with a particular topic embedding) by performing a nearest neighbor computation with items in the vocabulary. 
% This topic interpretation step also relies on our \compose\ function. 
Assume we have a vocabulary $V$ of words and phrases derived from the target collection (the phrase extraction process is detailed in appendix ~\ref{sec:phrase_extraction}). We pass each item in our vocabulary through \name\, which is trained to place words and phrases of variable length in the same semantically-meaningful vector space. The resulting vectors then form an embedding matrix $\bmat{L}$ of size $|V| \times d$ whose rows contain the corresponding output of the \name\ function. We can efficiently vectorize the topic interpretation by computing $\bmat{R}\bmat{L}^\mathsf{T}$, which results in a $K \times |V|$ matrix where each row corresponds to the inner products between a topic embedding and the vocabulary representations.

\subsection{Human evaluations on learned topics}

We compare \name\ against a slate of both neural and non-neural topic model baselines, including prior phrase-based topic models, on three datasets from different domains, using various topic sizes. Overall, we identify three key takeaways from our experiments: (1) \name\ produces more coherent topics than other phrase-based topic models and is competitive with word-level topic models; (2) \name's topics remain coherent even with large numbers of topics (e.g., 500-1000), unlike word-level models; and (3) despite the increase in vocabulary, \name's assignment of topics to documents is not impaired.

\subsubsection{Experiment Setup}
\label{sec:experiment-setup}
We experiment with three datasets (denoted as \textbf{Wiki}, \textbf{Story}, and \textbf{Reviews}) across three different domains (Wikipedia \citep{wk103}, fictional stories \citep{storium2020}, and online user product reviews \citep{he2016ups}). The datasets differ considerably in terms of document length and vocabulary. \footnote{Details of the datasets can be found in the Appendix \S \ref{sec:phrase_extraction}.}

We compare \pntm\ against four strong topic modeling baselines, two of which also incorporate phrases into topic interpretation (all neural models were trained on a single Nvidia RTX 2080Ti GPU in less than 3 hours): 

\paragraph{LDA:} LDA is a popular probabilistic generative model that learns document-topic and topic-word distributions \citep{blei2003latent}. We use Mallet \citep{mccallum2002mallet} with a parameter update every $20$ intervals for $1000$ total passes. We empirically observe that these hyperparameters produce the highest quality topics.

\paragraph{pLDA:} \citet{mimnophrase} incorporate phrases into LDA by simply converting them into unique word types (e.g., ``critically acclaimed'' to ``critically\_acclaimed'') and then run LDA as usual. We denote this method as \emph{phrase-LDA}  (pLDA). \footnote{We use the same phrase vocabulary used for our \pntm\ model (Table \ref{tab:corpusstats}) and the same hyperparameters as LDA.}

\paragraph{TNG:} We also compare our approach to the Mallet implementation of the topical $n$-gram model (TNG) of \citet{wang2007topical}, which learns to associate documents with topics while inducing a combined unigram and phrase vocabulary.

\paragraph{UNTM:} Finally, we implement a unigram version of \pntm, setting the word embedding function to simply average pretrained GloVe vectors. This model was also used by \citet{storium2020} and is based on the dictionary learning autoencoder originally proposed by \citet{rmn2016}.
\linebreak
\linebreak
Following \citet{chang2009reading}, we perform two different sets of human evaluation experiments on Amazon Mechanical Turk: (1) \textbf{word intrusion}, which measures topic coherence, and (2) \textbf{topic-to-document relatedness}, which evaluates whether a topic assigned to a document is actually relevant.
We implement the \textbf{word intrusion} task %which evaluates topic coherence (how related the most probable words associated with a topic are),
by giving crowd workers a list of six words or phrases and asking them to choose which one does not belong with the rest. The intruder is a highly-probable word or phrase sampled from a \emph{different} topic. We evaluate topic coherence through the \emph{model precision} metric \cite{chang2009reading}, which is simply the fraction of judgments for which the crowd worker correctly chose the intruder, averaged over all of the topics in the model. Similarly, for the \textbf{topic-to-document relatedness} task, we present crowdworkers with a passage from Wikipedia and two topics from the model, one of which is the most probable topic assigned by the model. Then, we ask them to choose which topic best matches the passage and report the fraction of workers agreeing with the model.\footnote{For all crowd experiments, we obtain three judgments per example, and we only allow qualified workers from English-speaking countries to participate. We restrict our tasks to workers with at least 97\% HIT approval rate and more than 1,000 HITs approved on Mechanical Turk. Workers are paid \$0.07 per judgment, which we estimate is a roughly \$10-12 hourly wage.}

\subsubsection{Topic modeling results}
\paragraph{\pntm\ produces more coherent topics than other phrase-based topic models.} 
Table \ref{tab:word-intrusion} contains the results of the word intrusion task run over all three datasets. In these experiments, we set the number of topics $K$ to 50 for all models and use the same vocabulary for pLDA and \pntm\ (as TNG induces phrases, we cannot control for its vocabulary). Compared to the other phrase-based models (TNG and pLDA), \pntm\ achieves substantially higher model precision. Notably, both neural topic models achieve higher model precision than LDA-based counterparts, and UNTM yields slightly more coherent topics than \pntm.\footnote{Two or more workers agree on the same choice $90.5\%$ of the time, indicating high degree of agreement; more details are in Appendix Table \ref{tab:word-intrusion-agree}}

\begin{table}[t]
    \centering\small
\begin{tabular}{lrrrr}
\toprule
  \bf Model & \bf Wiki & \bf Story & \bf Reviews & \bf Average\\
  \midrule

%   \pntm & 75.8
%     & 72.1
%     & 62.3
%     & 70.1 \\

  \pntm & \textbf{83.3}
    & \textbf{76.7}
    & \textbf{77.3}
    & \textbf{79.1} \\

  pLDA & 55.3
    & 48.7
    & 50.7
    & 51.6 \\ 

  TNG & 37.3
    & 58.7
    & 60.0
    & 52.1 \\\midrule
    
  UNTM & 76.9
    & 70.0
    & 62.0
    & 69.6 \\ 

  LDA & 48.7
    & 48.0
    & 52.7
    & 49.8 \\

  \bottomrule
\end{tabular}
    \caption{\pntm\ achieves higher model precision on word intrusion than other phrase-based topic models (top), and its topics are of similar quality to word-level neural topic models (UNTM).}
    \label{tab:word-intrusion}
\end{table}

\begin{figure}[t]
    \centering
  \includegraphics[width=1.0\linewidth, height=0.6\linewidth]{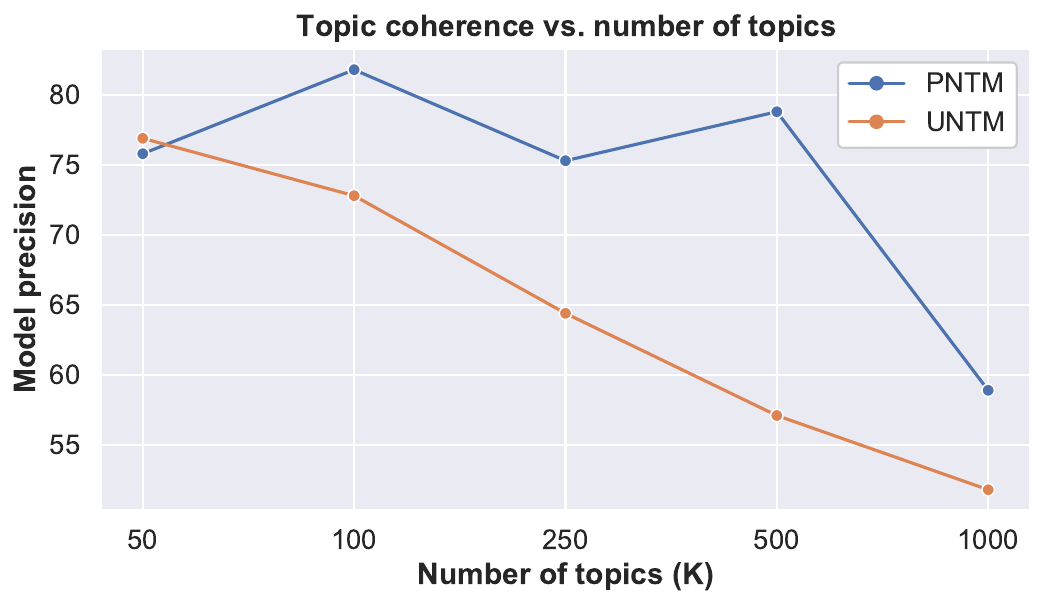}
  \caption{Word intrusion experiments show that \pntm\ achieves higher model precision (y-axis) than UNTM when the number of topics ($K$) is increased.}
  \label{fig:prec_vs_k}
\end{figure}

\paragraph{\pntm\ maintains high topic coherence when trained with more topics.}
One conceivable advantage of \pntm\ is that incorporating phrases into its vocabulary allows it to model topics at a finer granularity than unigram models. To test this hypothesis, we compare \pntm\ to its unigram-level neural counterpart UNTM across different values of $K$ ranging from 50 to 1000. Figure \ref{fig:prec_vs_k} plots the model precision derived from word intrusion experiments from both systems on Wikipedia, and Table \ref{tab:topicdump} includes six topics sampled from the $K=1000$ model. While model precision is similar between the two models when $K=50$, \pntm\ produces higher quality topics as $K$ increases. This increase in topic quality in \pntm\ signals that incorporating phrases into topic descriptions enable more topics to capture coherent and meaningful information.

\paragraph{\pntm\ is competitive with existing topic models on topic-to-document relatedness.} Table \ref{tab:topic-intrusion} shows that all five models achieve similar results with TNG performing slightly worse than the rest. A worker accuracy of close to 90\% signals that \pntm\ topics are assigned to relevant documents. Overall, \pntm\ is the only phrase-based model to achieve high scores in both word intrusion and topic-to-document relatedness tasks, showing that it learns higher quality topics without sacrificing relevance. The Krippendorf's $\alpha$ in Table \ref{tab:topic-intrusion} is a reliability statistic that measures inter-annotator agreement.

\paragraph{\pntm\ exhibits topic correspondence between related datasets.} We observe that the topics induced by \pntm\ on different but related datasets have correspondences when trained using the same random seed. More concretely, each topic in \pntm\ is associated with an index (denoting the corresponding row of the \bmat{R} matrix), and we observe correspondences between topics with the same index trained on different datasets.  Table~\ref{tab:topic-bijec} contains examples of this phenomenon; for instance, a topic on nightlife from a model trained on the ``fantasy classic'' Storium genre contains \emph{bar, drinking, the tavern}, while the topic with the same index from a model trained on ``occult pulp horror'' stories contains \emph{nightclub, clubbing, partygoer}. This ability provides practitioners with potentially new ways of exploring and comparing different collections of text, and it is not something easily implemented within LDA-based models. We theorize that such correspondences are possible because the learned topic embeddings do not move far away from their random initializations, which could be an effect of the orthogonality regularization.\footnote{The average L2 distance between the learned topic vectors and their random initializations is 2.72, while the average L2 distance amongst the learned topic vectors themselves is 3.65. }

\begin{table}[t]
	\begin{center}
		\small
		\begin{tabular}{ lp{4.5cm}}
			\toprule
			
			 \textbf{Genre} & \textbf{Corresponding topics} \\
			\midrule
			
			Fantasy Classic & curious, scanning, surveying, his vision, being watched \\
			Space Adventure & sensor, inspect, monitoring, check it out, spectrographic \\
			\midrule
            Fantasy Classic & bar, drinking, tavern, bartender, the tavern \\
            Pulp Horror & nightclub, clubbing, partygoer, nightlife, his drink\\
			\bottomrule
			
% % 			 & \textbf{Dataset} & \textbf{Topic Descriptions by Nearest Neighbour} \\
% % 			\cmidrule{2-3}
			
% 			Topic on \newline ``Positive Reviews" from the \textbf{Reviews} corpus \newline 
% 			& 
% 			Books \newline 
% 			CDs and Vinyl \newline \newline
% 			Electronics \newline
% 			Movies and TV
			
% 			& 
% 			loved it, love this book, my favorite book, love the book, my all - time favorites \newline 
% 			my favorite song, my favorite tracks, one of my favorite albums, one of my favorite bands, my mp3 player\newline 
% 			buying, bought it, worth buying, buying it, a great purchase\newline
% 			liked it, my favourite, a favorite of mine,  kept me interested, held my interest \\
			
% 			\midrule
			
% % 			 & \textbf{Subset} & \textbf{Topic Descriptions by Nearest Neighbour} \\
% % 			\cmidrule{2-3}
			
% 			Topic on \newline ``Negative reviews" from the \textbf{Reviews} corpus \newline 			& 
% 			Books \newline 
% 			CDs and Vinyl \newline 
% 			Electronics \newline
% 			Movies and TV
			
% 			& 
% 			boring, bland, worthless, poorly written, the worst book \newline 
% 			annoying, disappointment, the lack, boredom, pessimist\newline 
% 			required reading, read in years, read this review, read in a while, read the whole thing\newline
% 			miserable, very boring, no money, bad movies, are boring\\
            
% 			\bottomrule
	
		\end{tabular}
	\end{center}
	\caption{Examples of topic correspondences between models trained on different genres of stories. Matching topics have the same index in the \bmat{R} matrix.}
	\label{tab:topic-bijec}
\end{table}

\pntm\ also exhibits other useful properties such as the ability to interpret topics with phrases of various length (including even sentences). Qualitative examples of these phenomena are provided in Appendix \S \ref{sec:pntm-qualitative}

\begin{table}[t]
    \centering\small
\begin{tabular}{crc}
\toprule
  \bf Model & \bf Accuracy & \bf Krippendorf's $\alpha$  \\
  \midrule

%   \pntm & 86.8 & 0.7616 \\

  \pntm & 89.3 & 0.7084 \\ % rerun for camera-ready

  pLDA & 89.3 & 0.6259 \\ 

  TNG & 78.7 & 0.4876 \\\midrule
    
  UNTM & 80.7 & 0.7981 \\ 

  LDA & 90.0 & 0.7084\\

  \bottomrule
\end{tabular}
    \caption{Results of our topic-to-document relatedness experiments. \pntm\ achieves competitive worker accuracy to all other models, which indicates that its topic distribution assignments are not harmed by the inclusion of phrases to its vocabulary.}
    \label{tab:topic-intrusion}
\end{table}

\section{Conclusion}

We propose \name, which induces powerful phrase embeddings by fine-tuning BERT with two contrastive objectives on datasets of lexically diverse phrase-level paraphrases and phrases-in-context. \name\ consistently outperforms strong baseline models on a suite of phrasal semantic relatedness tasks, even when lexical overlap cues are removed. These results suggest that \name\ looks beyond simple lexical overlap to capture complex phrase semantics. Finally, we integrate \name\ into a neural topic model to enable phrase-based topic interpretation, and show that the resulting topics are more coherent and meaningful than competing methods.
% \newpage
\section*{Ethical considerations}

For all datasets and experiments, we use publicly available datasets from sources such as Wikipedia, Storium stories, and Amazon public reviews. We respect the privacy of all data contributors. In all crowdsourced evaluations, we strive to pay Mechanical Turkers with competitive payments.

We modify BERT embeddings in this project. Pretrained language models such as BERT are known to produce embeddings that raise ethical concerns such as gender \citep{gender-bias} and racial biases \cite{football2019, bommasani-etal-2020-interpreting}, and can also output other offensive text content. Practitioners may consider employing a post-processing step to filter out potentially offensive content before releasing the final output.
\section*{Acknowledgments}

We are grateful to Kalpesh Krishna, Xiang Lorraine Li, Varun Manjunatha, Simeng Sun, Zhichao Yang, and to the UMass NLP group for many helpful discussions.  We would also like to thank He He for recommending the use of PAWS as evaluation data, Tu Vu for the valuable suggestions on paper writing, and the anonymous reviewers for their insightful feedback. SW and MI were supported by award IIS-1955567 from the National Science Foundation (NSF).

% \newpage

\bibliography{bib/emnlp2021}
\bibliographystyle{acl_natbib}

\newpage
\appendix
\newpage
\clearpage
\newpage
\section{Appendix}
\label{sec:appendix}

\subsection{Source phrase extraction from Books3 Corpus}
\label{sec:source_phrase_extraction}

The Books3 Corpus \citep{pile} is a huge-scale collection of books from a variety of genres. We mine the Books3 to extract phrases by selecting constituency chunks. Particularly, we use the fast Stanford shift-reduce parser\footnote{https://nlp.stanford.edu/software/srparser.html} from~\citet{corenlp}, collecting all verb, noun, adjective, and adverb phrases from the data and keep the top 100K phrases with the highest frequency. We do not keep prepositional phrases as we find high overlap between prepositional phrases with noun phrases empirically.

\subsection{Phrase vocabulary extraction from dataset}
\label{sec:phrase_extraction}
Given the Wikipedia corpus \citep{wk103}, we first include all word types detected by spaCy's English tokenizer \citep{spacy} that occur more than five times in the corpus. We augment this vocabulary with phrases by extracting constituent chunks from the output of a constituency parser. We use the the same shift-reduce parser as in appendix Section \ref{sec:source_phrase_extraction}. Specifically, we extract all verb, noun, adjective, and adverb phrases from the data, and add the most frequent 75K phrases into our \bmat{L} matrix for topic interpretation as in Section \ref{sec:intepret-topics}. We omit prepositional phrases as they overlapped significantly with noun phrases. We perform the same vocabulary creation steps for the other two datastest (\textbf{Story} and \textbf{Reviews}) to extract all datasets (Table \ref{tab:corpusstats})

\begin{table}[t]
    \centering\small
\begin{adjustbox}{max width=0.45\textwidth}

\begin{tabular}{lrrrr}
\toprule
  \bf Dataset & \bf \# Docs & \bf \# Words & \bf \# Phrases & \bf Tok/doc\\
  \midrule

  Wikipedia & 304K
    & 47.2K
    & 75K
    & 396 \\

  Storium & 419K
    & 44.0K
    & 75K
    & 190 \\ 

  Reviews & 10K
    & 32.4K
    & 75K
    & 101 \\
  \bottomrule
\end{tabular}
\end{adjustbox}

    \caption{Corpus statistics for our three datasets, including the number of unique word and phrase types in our precomputed vocabulary. Note that we cap the number of unique phrases to the 75K most frequent.}
    \label{tab:corpusstats}

\end{table}

\subsection{Agreement among crowdsourced workers}
Evaluations from crowdsourced human evaluations show high inter-annotator agreement, indicated by close to 90\% of 2 or more workers agreeing on the same choice.

\begin{table}[h]
    \centering
\begin{tabular}{crrrr}
\toprule
  Model & Wiki & Story & Reviews \\
  \midrule

  \pntm & 92.0
    & 90.0
    & 96.0 \\

  pLDA & 96.0
    & 94.0
    & 82.0 \\ 

  TNG & 92.0
    & 94.0
    & 96.0 \\\midrule
    
  UNTM & 86.0
    & 96.0
    & 90.0 \\ 

  LDA & 88.0
    & 88.0
    & 90.0 \\

  \bottomrule
\end{tabular}
    \caption{Annotator agreement statistics (percentage of questions where two or more workers agree on the same answer) of our word intrusion experiments across datasets and models.}
    \label{tab:word-intrusion-agree}
\end{table}

\subsection{Qualitative Evaluation on \pntm: Interpreting topics with sentences}
\label{sec:pntm-qualitative}

% \subsubsection{Topic correspondence}
% We observe that the topics induced by \pntm\ on different but related datasets exhibit correspondences when trained using the same random seed. More concretely, each topic in \pntm\ is associated with an index (denoting the corresponding row of the \bmat{R} matrix), and we observe correspondences between topics with the same index trained on different datasets.  Table~\ref{tab:topic-bijec} contains examples of this phenomenon; for instance, a topic on nightlife from a model trained on the ``fantasy classic'' Storium genre contains \texttt{\{bar, drinking, the tavern\}}, while the topic with the same index from a model trained on ``occult pulp horror'' stories contains \texttt{\{nightclub, clubbing, partygoer\}}. 

% This ability provides practitioners with potentially new ways of exploring and comparing different collections of text, and it is not something easily implemented within LDA-based models. We theorize that such correspondences are possible because the learned topic embeddings do not move far away from their random initializations, which could be an effect of the orthogonality regularization.\footnote{The average L2 distance between the learned topic vectors and their random initializations is 2.72, while the average L2 distance amongst the learned topic vectors themselves is 3.65. }

% \input{tables/table-topic-bijec}

% \subsubsection{Interpreting topics with sentences}
% \paragraph{Interpreting topics with sentences}

Another capability that sets \pntm\ apart from existing models is \emph{sentence-level} topic interpretation, which offers an even more fine-grained understanding of learned topics. This functionality has potential to help with automatic topic labeling, which traditionally has been a manual process because the most probable words in a topic are not necessarily the most descriptive words of a particular high-level theme. Since the underlying BERT model of \pntm's embedding function is fine-tuned on both sentence and phrase-level data, its representations are semantically meaningful across multiple scales of text. We also do not have to retrain the model to interpret topics with sentences; rather, we just have to encode the training sentences (or potentially sentences from an external corpus) with our embedding model (\pntm) and then add them to the vocabulary (i.e., as additional rows in the $\bmat{L}$ matrix).

\begin{table}[t]
	\begin{center}
		\footnotesize
		\begin{tabular}{p{7cm} }
			\toprule

            \multicolumn{1}{c}{\emph{Interpreting with words / phrases}}\\
            missourian, american history, county route, alabama, confederate, a state highway \\
            \midrule
            \multicolumn{1}{c}{\emph{Interpreting with sentences}}\\

            \emph{1.} At its 1864 convention , the Republican Party selected Johnson , a War Democrat from the Southern state of Tennessee , as his running mate .
            
            \vspace{0.1cm}
            \emph{2.} Burnett also raised a Confederate regiment at Hopkinsville , Kentucky , and briefly served in the Confederate States Army .
            
            \vspace{0.1cm}
             \emph{3.} Parker was nominated for Missouri 's 7th congressional district on September 13 , 1870 , backed by the Radical faction of the Republican party .\\
			\bottomrule
		\end{tabular}
	\end{center}
	\caption{Sentence-level interpretation makes it clear that this topic is about Civil War-era American history, while word and phrase interpretation offers a more high-level view.}
	\label{tab:sent-rep}
\end{table}

Table~\ref{tab:sent-rep} contains one such example, which is a topic from a \pntm\ model trained with $K=50$ on Wikipedia. When interpreted with just words and phrases, the topic looks like it focuses on Southern and Midwestern U.S. states and their history. However, when interpreting the same topic with sentences from the training set, we observe that the most probable sentences for this topic all reference the Civil War / Reconstruction era of U.S. history. These kinds of observations might influence not only a practitioner's labeling of a particular topic, but also how they use the topic model itself.

\subsection{Topics from \pntm\ with different embedding functions}
\label{sec:topicdump-qualitative}
We present topic samples from three versions of \pntm, using BERT, SpanBERT, and \name\ respectively as the embedding function. Other than the embedding function used, the three topic models have the same architecture and are trained with the same hyperparameters. The training dataset is \textbf{Wiki} (the same dataset in Section \ref{sec:experiment-setup}), with the number of training epochs $=300$ and the number of topics $=50$.

Qualitatively, we observe that the \name-based topic model produces the highest quality topics that are both lexically diverse and also coherent, in Table \ref{tab:phrasebert-topic-dump}. The topics from the BERT-based topic model (Table \ref{tab:bert-topic-dump}) have lower quality as the over-reliance on word content overlap makes some topics less informative (e.g., ``his album", ``the album", ``an album" ...). However, the topic descriptions are largely interpretable as the words and phrases used are still semantically coherently. The SpanBERT-based topic model, on the other hand, produces even lower quality topics as  the topic descriptions are incoherent in many cases, as shown in Table \ref{tab:spanbert-topic-dump}.

\begin{table}
	\begin{center}
		\small
		\begin{tabular}{ p{7cm}}
			\toprule
			
			%1 
            winning,
            semifinalist,
            finisher,
            a race,
            raceme,
            race,
            the race,
            the race 's,
            side rowing competition,
            formula one\\ \midrule
            
            %2
            bullfighter,
            bullfighting,
            showman,
            wwe 's,
            wwe smackdown,
            wwe day,
            wrestle,
            wrestler,
            the wwe championship,
            wrestling\\ \midrule
            
            %3
            the gatehouse,
            the plant,
            the farm,
            the estate,
            the building,
            the fort,
            the castle,
            landscaping,
            was built,
            the monument \\\midrule
            
            %4
            the beatles,
            his album,
            the album 's,
            discography,
            the album,
            an album,
            beatles,
            this album,
            the beatles ',
            their album\\ 
            \midrule
            
            %5
            tropical cyclones,
            a tropical depression,
            developed into a tropical depression,
            a tropical storm warning,
            a tropical cyclone,
            tropical storm arlene,
            tropical storm status,
            tropical storm,
            the tropical storm,
            a tropical storm\\ \midrule
            
            %6
            a mother,
            her parents,
            his parents,
            her father,
            her father 's,
            her mother 's,
            his mother,
            her mother,
            his mother 's,
            her parents ' \\
			\bottomrule

		\end{tabular}
	\end{center}
	\caption{A sample of six topics induced by \pntm\ with BERT as the embedding model}
	\label{tab:bert-topic-dump}
\end{table}

\begin{table}
	\begin{center}
		\small
		\begin{tabular}{ p{7cm}}
			\toprule
			
			%1
            two episodes,
            expressible,
            side rowing race,
            tourmaline,
            followed throughout the united kingdom,
            the island 's,
            drive,
            sidecar,
            flywheel,
            cockpit \\ \midrule
            
            %2
            lieutenant colonel,
            new mexico,
            midshipman,
            postmenopausal,
            generalship,
            generalissimo,
            ambassadorship,
            valedictorian,
            the wwe championship,
            the spanish \u2013 american war\\ \midrule
            
            %3
            ellipse,
            opulent,
            meetinghouse,
            embark,
            rapidity,
            swiftly,
            the 13th century,
            institution,
            rapidly,
            gradual\\\midrule
            
            %4
            songbook,
            the novel,
            a novel,
            this book,
            her book,
            storybook,
            fiction book,
            novelette,
            novelization,
            novelisation\\ 
            \midrule
            
            %5
            major intersections,
            rainstorm,
            thunderstorm,
            torrential,
            a storm,
            the race 's,
            cloudy,
            high winds,
            major hurricanes,
            windstorm\\ \midrule
            
            %6
            satanic,
            luciferin,
            lynchpin,
            judas,
            blackmailer,
            kidnapping,
            satanist,
            bosch,
            vaulting,
            afire\\
			\bottomrule

		\end{tabular}
	\end{center}
	\caption{A sample of six topics induced by \pntm\ with SpanBERT as the embedding model}
	\label{tab:spanbert-topic-dump}
\end{table}

\begin{table}
	\begin{center}
		\small
		\begin{tabular}{ p{7cm}}
			\toprule
			
			%1
            the semifinal,
            olympic,
            marathon,
            raceme,
            bicyclist,
            semifinalist,
            side rowing race,
            racer,
            place finish,
            side rowing competition\\ \midrule
            
            %2
            powerful,
            wrestler,
            the forces,
            most powerful,
            demonic,
            the organization,
            a force,
            power,
            an organization,
            dark forces\\ \midrule
            
            %3
            newly built,
            terrace,
            atrium,
            architecture,
            foyer,
            the building,
            the city centre,
            facade,
            architecturally\\\midrule
            
            %4
            musician,
            his music,
            musical,
            concerto,
            chorale,
            live performances,
            a concert,
            accompaniment,
            pianistic,
            antiphonal\\ 
            \midrule
            
            %5
            tropical depression,
            a tropical cyclone,
            a category 2 hurricane,
            a tropical storm,
            a category 1 hurricane,
            tropical storm status,
            tropical storm,
            a tropical disturbance,
            developed into a tropical depression,
            a tropical depression\\ \midrule
            
            %6
            a police officer,
            criminology,
            criminalisation,
            criminal cases,
            illegality,
            law enforcement,
            criminalization,
            felony,
            criminality,
            misdemeanor\\
			\bottomrule

		\end{tabular}
	\end{center}
	\caption{A sample of six topics induced by \pntm\ with \name\ as the embedding model}
	\label{tab:phrasebert-topic-dump}
\end{table}

\end{document}